# SA-NET.V2: REAL-TIME VEHICLE DETECTION FROM OBLIQUE UAV IMAGES WITH USE OF UNCERTAINTY ESTIMATION IN DEEP META-LEARNING

M. Khoshboresh-Masouleh , R. Shah-Hosseini*

School of Surveying and Geospatial Engineering, College of Engineering, University of Tehran, Tehran, Iran – (m.khoshboresh, rshahosseini)@ut.ac.ir

**ABSTRACT:**

In recent years, unmanned aerial vehicle (UAV) imaging is a suitable solution for real-time monitoring different vehicles on the urban scale. Real-time vehicle detection with the use of uncertainty estimation in deep meta-learning for the portable platforms (e.g., UAV) potentially improves video understanding in real-world applications with a small training dataset, while many vehicle monitoring approaches appear to understand single-time detection with a big training dataset. The purpose of real-time vehicle detection from oblique UAV images is to locate the vehicle on the time series UAV images by using semantic segmentation. Real-time vehicle detection is more difficult due to the variety of depth and scale vehicles in oblique view UAV images. Motivated by these facts, in this manuscript, we consider the problem of real-time vehicle detection for oblique UAV images based on a small training dataset and deep meta-learning. The proposed architecture, called SA-Net.v2, is a developed method based on the SA-CNN for real-time vehicle detection by reformulating the squeeze-and-attention mechanism. The SA-Net.v2 is composed of two components, including the squeeze-and-attention function that extracts the high-level feature based on a small training dataset, and the gated CNN. For the real-time vehicle detection scenario, we test our model on the UAVid dataset. UAVid is a time series oblique UAV images dataset consisting of 30 video sequences. We examine the proposed method's applicability for stand real-time vehicle detection in urban environments using time series UAV images. The experiments show that the SA-Net.v2 achieves promising performance in time series oblique UAV images.

## 1. INTRODUCTION

The development of remote imaging platforms has induced the researchers to use these platforms in many real-world applications where is important real-time object detection, and robust monitoring method (Giordan et al. 2020; Kiribayashi, Yakushigawa, and Nagatani 2018). A drone or an Unmanned Aerial Vehicle (UAV) can be equipped with small and compact imaging and positioning sensors, such as optical camera, thermal sensor, Light Detection And Ranging (LiDAR), and Global Navigation Satellite System (GNSS) receiver for generating a wide range of geospatial data and details in a short time for the study of natural objects and environmental monitoring, with very low operating costs (Svedin et al. 2021; Mustafah, Azman, and Akbar 2012). In this regard, the variety and multimodality data from Earth can impose difficulty on both datasets generating and processing methods since it is hard to find the right strategy that matches their learning preferences. In recent years, UAV imaging is a suitable solution for real-time monitoring different vehicles on the urban scale. Real-time vehicle detection with the use of uncertainty estimation in deep meta-learning for the portable platforms potentially improves video understanding in real-world applications with a small training dataset, while many vehicle monitoring approaches appear to understand single-time detection with a big training dataset.

Deep meta-learning is an inductive transfer system whose main goal is to improve generalization ability for multiple tasks (Huisman, van Rijn, and Plaat 2021). The investigation of previous methods proves that there are still many important problems, such as robustness ability for meta-learning from UAV images, that have not yet been adequately considered in relevant methods. The lack of datasets and algorithms for deep meta-learning especially in real-world applications is the main motivation of this study.

The purpose of real-time vehicle detection from oblique UAV images is to locate the vehicle on the time series UAV images by using semantic segmentation. Real-time vehicle detection is more difficult due to the variety of depth and scale vehicles in oblique view UAV images (Xie et al. 2018; Lyu et al. 2020). Motivated by these facts, in this manuscript, we consider the problem of real-time vehicle detection for oblique UAV images based on a small training dataset and deep meta-learning. The proposed architecture, called SA-Net.v2, is a developed method based on the SA-CNN (Khoshboresh-Masouleh and Shah-Hosseini 2021b) for real-time vehicle detection by reformulating the squeeze-and-attention mechanism. The SA-Net.v2 is composed of two components, including the squeeze-and-attention function that extracts the high-level feature based on a small training dataset, the gated CNN. For the real-time vehicle detection scenario, we test our model on the UAVid dataset. UAVid is a time series oblique UAV images dataset consisting of 30 video sequences (Lyu et al. 2020). We examine the proposed method's applicability for stand real-time vehicle detection in urban environments using time series UAV images. The main idea is to collect all the possible datasets and algorithms for deep meta-learning from UAV imagery sensors that exist until the writing of this research and use the efficient methods to scene understanding for real-world applications, such as vehicle monitoring. This study summarizes the novel methods of meta-learning and its research progress and real-world applications in meta-learning from UAV imagery sensors introduces the current main challenges in data mining and its development of related datasets and focuses on the analysis and elaboration of the research status of meta-learning in UAV imaging.

---

## 2. A COMPARISON BETWEEN TRANSFER AND MULTI-TASK LEARNING

Transfer learning is a paradigm to train on one task and transfer to a new task (Torrey and Shavlik 2010). Transfer learning consists of the transfer of knowledge from an old task to a newer task, which has some similarities with the old (Kulkarni and Nair 2021). Transfer and multi-task learning algorithms are not the same. Multi-task learning tries to learn the new and old task simultaneously, while transfer learning only aims at achieving high performance in the new task by transferring knowledge from the old task (M. R. Bayanlou and Khoshboresh-Masouleh 2021), as shown in Fig. 1. Figure 1 shows the different structures in transfer and multi-task learning for a drone image from an urban area.

In a transfer learning algorithm, a task based on a probabilistic viewpoint is defined as follows (Zhuang et al. 2020):

$$T = \{Y, P(Y|X)\} \quad (1)$$

where X denotes a feature space, Y denotes a label space, P(Y|X) denotes a predictive function based on the training dataset, and P(X) denotes a marginal probability distribution.

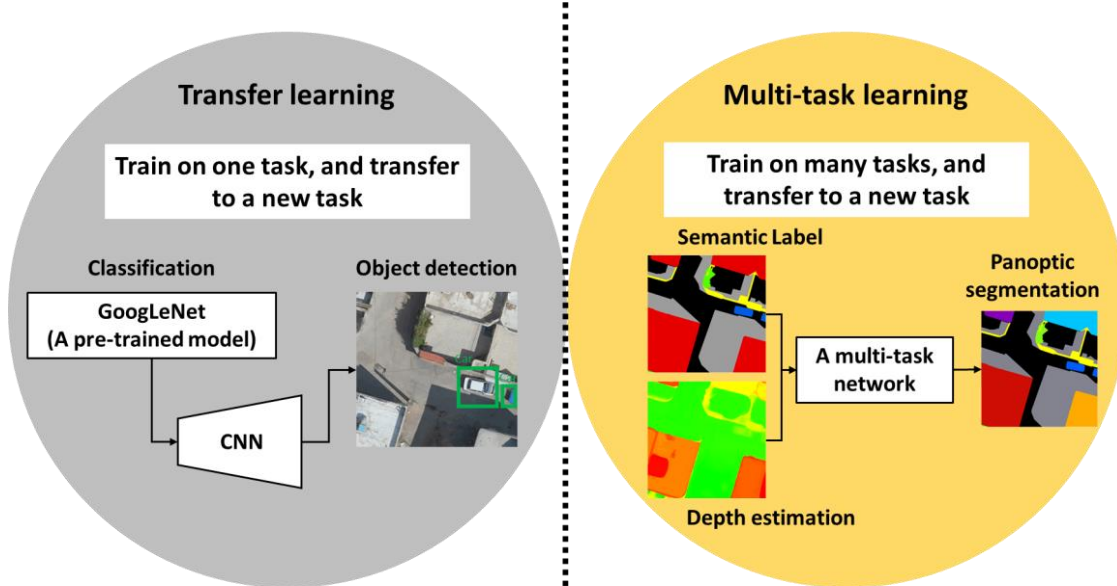

Figure 1. A comparison between transfer and multi-task learning

As a result, transfer learning aims to enhance the conditional probability distribution in a new task based on an old task, where the new task is against the old task. To review the transfer learning topic for the labeled and unlabeled dataset, the existing problems can be broadly categorized into homogeneous transfer learning and heterogeneous transfer learning. The homogeneous transfer learning approaches develop rulesets to correct both the marginal and conditional distribution differences in the old and new tasks. Heterogeneous transfer learning approaches bridge the gap between feature spaces and reduce the problem to a homogeneous transfer learning issue where marginal or conditional distribution differences will need to be corrected (Zhuang et al. 2020).

## 3. FEW-SHOT LEARNING FOR UAV IMAGE UNDERSTANDING

To learn from a limited number of training samples with labeled data, a new strategy called few-shot learning is proposed. In real-world applications, annotating images or videos is very expensive. To build effective machine learning models in these applications, deep few-shot learning methods have been developed and prove to be a robust approach in small training data (Feng et al. 2021; Voulodimos et al. 2021).

Few-shot learning can help relieve the burden of collecting large-scale supervised data. For example, although a pre-trained model, e.g., DeepMultiFuse (Khoshboresh-Masouleh and Akhoondzadeh 2021), outperforms humans on multispectral UAV images, each class needs to have sufficient labeled images which can be laborious to collect.

## 4. SM DATASET

The SAMA-VTOL dataset, also known as SM, (M. R. Bayanlou and Khoshboresh-Masouleh 2021; M. R. R. Bayanlou and Khoshboresh-Masouleh 2020) is composed of much larger multi-task information and with more data complexity in terms of the number of targets, which makes this dataset more adequate for multi-task learning for scene understanding from drone images.

There are 10 classes defined for the semantic segmentation task, including water, ground, parcel boundary, waste object, vehicle, farmland, vegetation, building shadow, building, and vegetation. SM dataset provides a multi-task dataset targeting semantic, panoptic, and depth labeling for urban scene analysis from fixed-wing drone images.

SM dataset is a new drone-based image dataset for a wide range of scientific projects in civil engineering, such as 3D urban modeling, urban/rural mapping, and digital surface model. Drone images play a crucial step in providing geospatial datasets. This dataset consists of suburb scene images with a forward overlap of 80% between images and a side overlap of 60% between flight lines from part of Esfahan, Iran. The characteristics that make the SM dataset an excellent scientific dataset are:

(a) High ground sampling distance (GSD).
(b) Post Processing Kinematic system for improving the spatial accuracy without ground control points (GCPs).
(c) Different landscape types, including different types of roofs for commercial/residential buildings, and vegetation.

## 5. IND DATASET

The IND dataset (Khoshboresh-Masouleh and Shah-Hosseini 2021a) consists of high-altitude drone images labeled for multi-modal building segmentation with a period of 2 years that have significant land-use changes from 8 various areas that sit in different cities in Indiana of the United States. The fully labeled IND contains a total of 300 multi-modal building instances.

The IND dataset includes major cities in Indiana images targeting building semantic labeling for the urban scenes. The IND dataset is 15 cm resolution and contains 256 training and 38 test aerial RGB-Normalized Difference Vegetation Index (NDVI)-Depth and pixel-level building labeled maps at 512×512 resolution, which brings new challenges, including shadows and occluded areas, vegetation covers, complex roofs, dense building areas, and large-scale variation.

## 6. UAVID DATASET

The UAVid dataset (Lyu et al. 2020) is a drone time-series dataset for image segmentation from urban scenes with an oblique view. There are 8 classes defined for this dataset, including building, road, tree, low vegetation, static car, moving car, human, and background.

The UAVid dataset consists of a time-series dataset targeting semantic labeling for urban scene analysis from an oblique drone perspective.

An overview of existing datasets for multi-task learning can be found in Table 1.

| Reference | SM | IND | UAVid |
|---|---|---|---|
| Data source | drone | High altitude drone | drone |
| Open access | No | Yes | Yes |
| Type | Orthophoto | Orthophoto | Video |
| Depth map | Very high-resolution | High-resolution | No |
| NVDI | No | Yes | No |
| Texture distortion | Low | Low | - |
| Semantic annotation | Yes | Yes (just buildings) | Yes |
| Semantic classes | 8-10 | 2 | 8 |
| Panoptic annotation | Yes (just buildings) | No | No |
| Color-based 3D point cloud | Yes | No | No |
| Image size (pix) | 2000×2000 | 512×512 | 4096×2160 |
| GSD | 2.5cm | 15cm | - |

Table 1. List of datasets for multi-task learning in drone imaging sensors

## 7. SA-NET.V2

The SA-Net.v2 proposed in this manuscript aims to improve the precision and inference time for real-time vehicle detection. This method is organized based on the theory of SA-CNN. Our network is different from similar models in the feature extraction produced by CNN. The SA-Net.v2 architecture takes the HRNet.v2 as the backbone and uses a region-based temporal aggregation Monte Carlo dropout (Huang et al. 2018), which can further improve the uncertainty modeling to guide the real-time vehicle detection. HRNet.v2 model is a pair of encoders and decoders. The encoder is HRNetV2-W48 and the decoder is the one convolution module and interpolation.

The squeeze-and-attention function learns non-local spectral features and multi-scale spatial representations in time-series images and therefore overcomes the constraints of convolutional blocks and mask generation. The squeeze-and-attention function is defined as follows:

Output = Upsampling [ReLU(F(P(input)))]   (2)

The gated CNN was efficient in minimizing the unnecessary transmission of information by using a convolutional block for extracting optimized time-series images and overcoming the end-to-end problem posed in target detection. The gated CNN is defined as follows:

gate-cnn = [batch_normal(lower encoder features*convolutional layers)].[upsampling(upper encoder features* convolutional layers)]   (3)

## 8. RESULTS

In this manuscript, the Intersection-over-Union (IoU) used in evaluating real-time vehicle detection, which is formulated as:

IoU = (2* true positives) / (true positives+ false positive+ false negative)   (4)

Moreover, the entropy measure for uncertainty estimation I used. Table 2 shows the quantitative results for real-time vehicle detection from oblique UAV images. Figure 2 shows the real-time vehicle detection based on SA-Net.v2 from oblique UAV images.

| Time series samples | IoU | Entropy |
|---|---|---|
| Set1, n=10 | 79.2 | 0.31 |
| Set2, n=10 | 92.4 | 0.16 |
| Set3, n=10 | 94.8 | 0.11 |
| Set4, n=10 | 89.9 | 0.23 |
| Set5, n=10 | 88.1 | 0.21 |
| Set6, n=10 | 91.4 | 0.18 |
| Mean | 89.3 | 0.20 |

Table 2. real-time vehicle detection comparisons on different types of test scenes based on SA-Net.v2

## 9. CONCLUSIONS

The proposed method, called SA-Net.v2, is trained on a few-shot dataset to provide better real-time vehicle segmentation performance for oblique UAV images. The experimental results highlight the abilities of the SA-Net.v2 to vehicle segmentation that include various challenges with IoU score of 89.3, and an entropy score of 0.20.

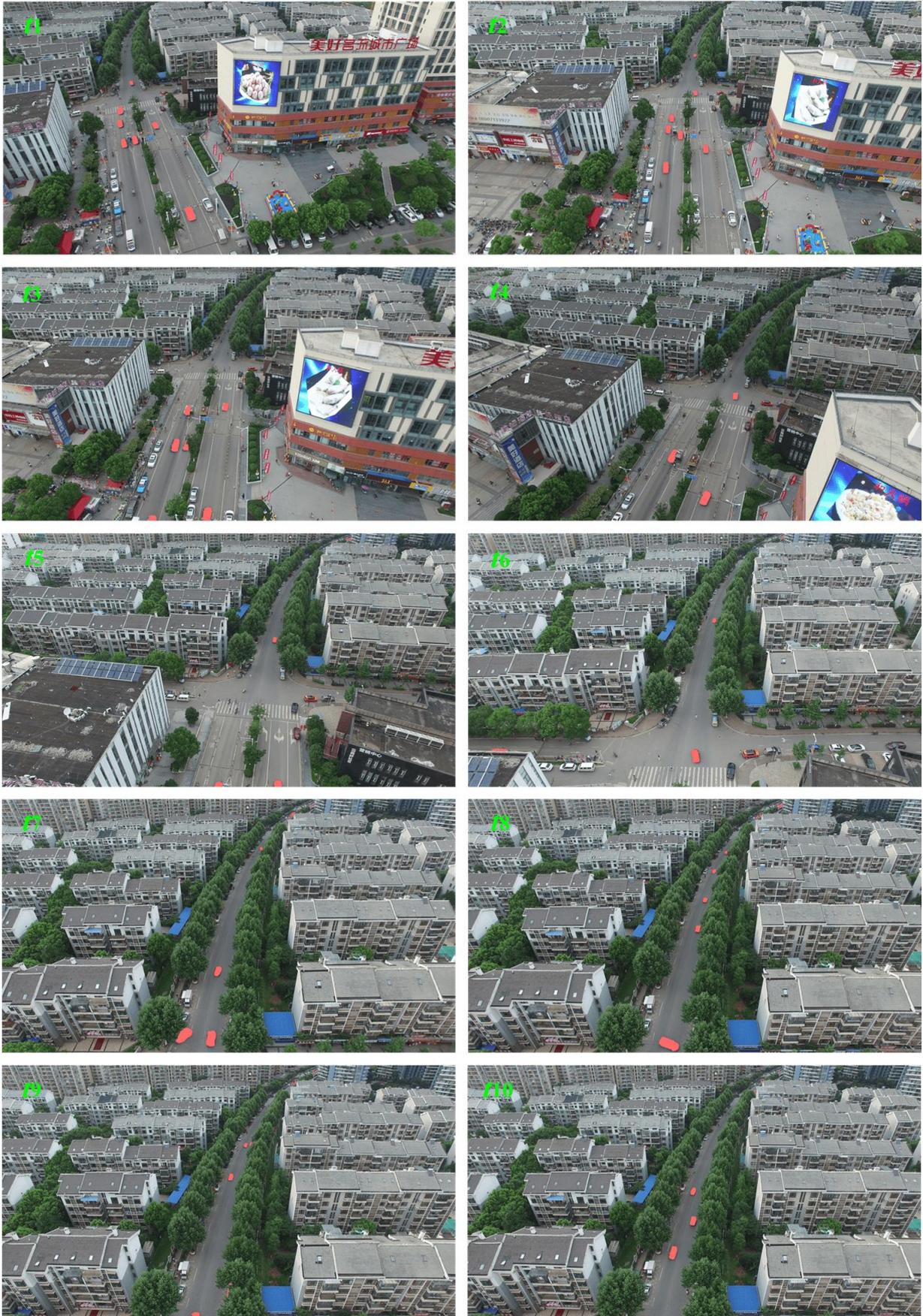

Figure 2. Example time-series vehicle detection from oblique UAV images. Red footprints indicate all vehicle instances